\documentclass[letterpaper]{article} 
\usepackage{aaai24}  
\usepackage{times}  
\usepackage{helvet}  
\usepackage{courier}  
\usepackage[hyphens]{url}  
\usepackage{graphicx} 
\usepackage{natbib}  
\usepackage{caption} 
\usepackage{algorithm}
\usepackage{algorithmic}
\usepackage{multirow}
\usepackage[table,xcdraw]{xcolor}
\usepackage{booktabs}
\usepackage{subfigure}
\usepackage{bm}
\usepackage{tcolorbox}
\usepackage{newfloat}
\usepackage{listings}
\DeclareCaptionStyle{ruled}{labelfont=normalfont,labelsep=colon,strut=off} 
\floatstyle{ruled}
\newfloat{listing}{tb}{lst}{}
\floatname{listing}{Listing}
\title{Coreference Graph Guidance for Mind-Map Generation}
\author{
    Zhuowei Zhang\textsuperscript{\rm 1},
    Mengting Hu\textsuperscript{\rm 1}\thanks{ Corresponding author},
    Yinhao Bai\textsuperscript{\rm 1},
    Zhen Zhang\textsuperscript{\rm 1}
}
\affiliations{
    \textsuperscript{\rm 1}College of Software, Nankai University

    \{zhuoweizhang, yinhao, zhangzhen\}@mail.nankai.edu.cn, mthu@nankai.edu.cn
}

\usepackage{bibentry}

\begin{document}

\maketitle

\begin{abstract}

Mind-map generation aims to process a document into a hierarchical structure to show its central idea and branches. Such a manner is more conducive to understanding the logic and semantics of the document than plain text. Recently, a state-of-the-art method encodes the sentences of a document sequentially and converts them to a relation graph via sequence-to-graph. Though this method is efficient to generate mind-maps in parallel, its mechanism focuses more on sequential features while hardly capturing structural information. Moreover, it's difficult to model long-range semantic relations. In this work, we propose a coreference-guided mind-map generation network (CMGN) to incorporate external structure knowledge. Specifically, we construct a coreference graph based on the coreference semantic relationship to introduce the graph structure information. Then we employ a coreference graph encoder to mine the potential governing relations between sentences. In order to exclude noise and better utilize the information of the coreference graph, we adopt a graph enhancement module in a contrastive learning manner. Experimental results demonstrate that our model outperforms all the existing methods. The case study further proves that our model can more accurately and concisely reveal the structure and semantics of a document. Code and data are available at \url{https://github.com/Cyno2232/CMGN}.

\end{abstract}

\begin{figure}
    \centering
    \includegraphics[width=0.48\textwidth]{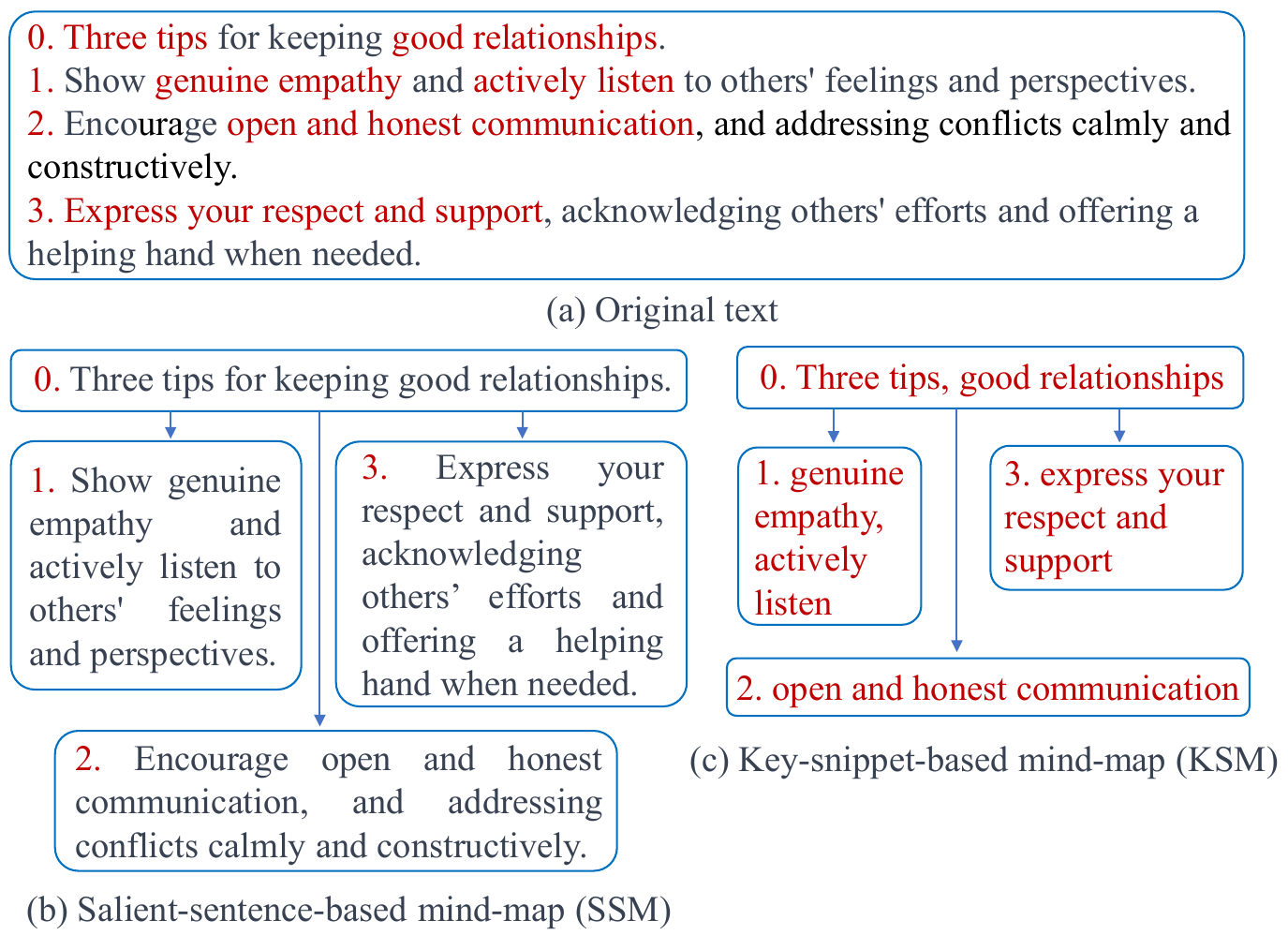}
\caption{A document (a) and two forms of mind-map derived from the document, namely salient-sentence-based mind-map (SSM) and key-snippet-based mind-map (KSM), constructed by sentences (b) or keywords (c).}

\end{figure}

\section{Introduction}

A mind-map generally consists of a series of hierarchical nodes, including central nodes and branches, which are conceptually connected and visually linked with lines \cite{kudelic2011mind,wei2019revealing,hu2021efficient}. Mind-map is designed to describe core concepts and to guide the text through a cognitive structure, which makes it easier to understand than reading the original text directly  \cite{dhindsa2011constructivist}. Figure 1 provides an example of a mind-map. As depicted, a node represents the idea of a single sentence, and an edge represents the governing relationships between sentences. There are generally two kinds of mind-map, namely salient-sentence-based mind-map (SSM) and key-snippet-based mind-map (KSM). The difference between them is that the nodes of SSM are constructed using whole sentences, while the nodes of KSM are constructed using keywords \cite{dhindsa2011constructivist}.

Many softwares have been invented to assist \emph{manual mind-map construction}, such as FreeMind, MindMapper, Visual Mind, etc \cite{kudelic2011mind}. Afterward, various approaches have been proposed to \emph{automatically generate mind-maps} from input texts. Some aim to identify the semantic relationship within one single sentence by using pre-established rules \cite{brucks2008assembling, rothenberger2008figuring} or syntactic parser \cite{elhoseiny2012english2mindmap}. Recent works \cite{wei2019revealing, hu2021efficient} generate mind-maps by detecting the semantic relations between different sentences in a document.

Concretely, the generation process is divided into two phases, as shown in Figure 2. Assume a document has $N$ sentences, the first phase aims to build a graph $\textbf{G}$ for all pairs of sentences. Each element in $\textbf{G}$ indicates the governing relationship between two sentences. Then, since the graph is redundant, the second phase prunes the extra edges to obtain the mind-map. Yet the number of sentence pairs increases exponentially with the length of the document, which raises the computational complexity. Recently, \citet{hu2021efficient} propose to improve the efficiency of the first phase by encoding sentences sequentially and feeding them into a sequence-to-graph module to obtain the relation graph. Although sequential encoding can raise efficiency, its perception of graph structure is limited. Additionally, as document length increases, it becomes more difficult for sequential encoders to capture semantic relationships over long distances. 

\begin{figure}
    \centering
    \includegraphics[width=0.48\textwidth]{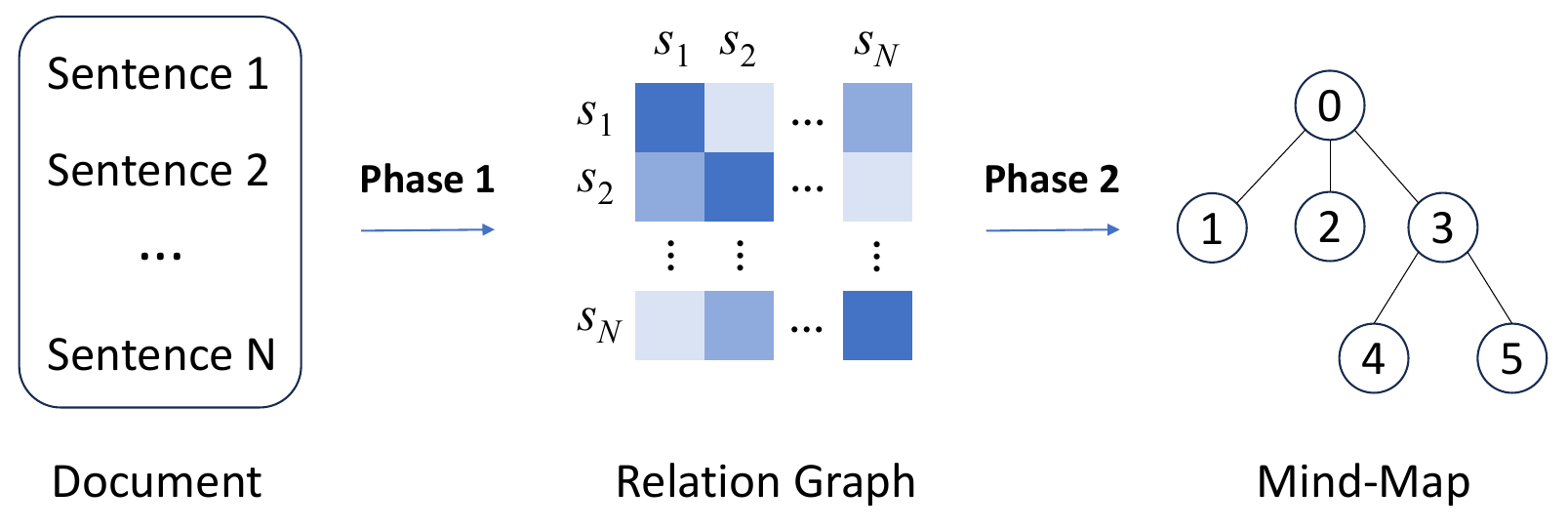}
\caption{Two-phases mind-map generation process. Phase 1 builds a relation graph for a given document. Phase 2 discards extra edges to yield the final mind-map.}

\end{figure}


To address the above issues, we propose a coreference-guided mind-map generation network (CMGN). Specifically, a coreference graph is first constructed to explicitly introduce prior graph structure information. The coreference graph contains the coreference semantic relationships, which imply the governing relation of the sentence at various positions in the document. Based on this graph, we employ a coreference graph encoder based on graph convolutional network (GCN) to further mine the potential semantic relationships between sentences, especially for long-distance sentence pairs. 

However, the semantic relations in coreference graphs are not always reliable. For instance, some coreference entities are unrelated to the subject of the document. To address the issue, we introduce the graph enhancement module, which is based on graph contrastive learning (GCL). Concretely, we utilize dual graph neural network (GNN) encoders with the same architecture, one of which is obtained by applying perturbations to all the parameters of the other. Then, we take original graph data as input and dual GNN encoders as counterparts to obtain two correlated views. With the encoder perturbation as noise, we can obtain two different embeddings for the same input as positive pairs, and take other graph data as negative pairs. We maximize the agreement of the positive pairs.

In summary, the main contributions of this paper are as follows.

\begin{itemize}

    \item We propose a novel mind-map generation method that constructs the coreference graph to involve graph structure information. Then we employ a graph encoder to capture long-range semantic relations.
    \item Since the information of the coreference graph is not always reliable, we employ a graph enhancement module to sufficiently utilize the information of the coreference graph.
    \item Extensive experimental results demonstrate
    that the proposed method outperforms the state-of-the-art method. Further analysis proves that the proposed method can generate more meaningful and concise mind-maps while maintaining efficiency.

\end{itemize}

\section{Related Works}

A mind-map is a diagram with a hierarchical structure, which can disclose the logical structure of a document \cite{buzan2006mind}. Of the two kinds of mind-map, SSM is similar to extractive text summarization \cite{zhong2020extractive}, which involves selecting and combining essential sentences or phrases from a given document to create a concise summary. Mind-map, by contrast, aims at not only the general idea of a document but also the relation of succession between paragraphs.

There are a number of works that use sentence-based graphs to generate text summarization. A previous study LexRank \cite{erkan2004lexrank} employs a graph-based approach to compute an adjacency matrix for sentence representation. This method relies on intra-sentence cosine similarity. However, generating a meaningful mind-map from this graph representation becomes challenging, as specific sentence pairs with semantic relations may possess zero lexical similarity. Additionally, several extractive summarization studies have utilized graph techniques. For instance, to enhance the ranking of sentences within a document, several methods have been proposed, including the utilization of bipartite graphs for sentence and entity nodes \cite{parveen2015integrating}, and weighted graphs featuring topic nodes \cite{parveen2015topical}. Recently, \citet{wang2020heterogeneous} utilize a heterogeneous graph for the purpose of capturing the relations between words and sentences. \citet{liu2021unsupervised} construct sentence graphs based on both the similarities and relative distances in the neighborhood of each sentence. However, these attempts of involving graph knowledge can hardly acquire governing relations between sentences, and thus can not reveal the logical structure of a document.

Graph contrastive learning (GCL) is a self-supervised learning algorithm for graph data. It learns to capture meaningful patterns and relationships within the graph by encouraging similar nodes or edges to be close in the embedding space while pushing dissimilar ones apart. \citet{sun2022contrastive} construct heterogeneous graphs from texts and expand the heterogeneous graph neural network model (HGAT) with simple neighbor contrastive learning. The negative samples are created by corrupting the edges of the graphs. \citet{hu-etal-2022-graph} build the graph through the entities and dependency tree of the given document. The negative samples are constructed by masking keywords, while the positive samples are constructed by masking non-keywords. \citet{xu2021contrastive} propose to construct a graph on top of the document passages to utilize multi-granularity information. Then they design a contrastive learning strategy where agreement among sub-document representations from the same document are maximized.
In the case of the above work, their negative samples were obtained by modifying the text or graph structure. Different from them, we directly perturb the parameters of the encoder and generate the negative samples from the perturbed encoder.

\section{Methodology}

\subsection{Problem Definition}
We define that a document $D$ consists of the sentences $\{s_1, s_2, ..., s_N\}$, and a sentence $s_i$ consists of the words $\{w_i^1, w_i^2, ..., w_i^L\}$, where $N$ and $L$ are the number of the sentences and the words respectively. The mind-map generation task can be defined as:
\begin{equation}
    D \rightarrow \mathbf{G} \rightarrow M
\end{equation}
where the input document $D$ is firstly converted to a relation graph $\mathbf{G}$. Then the final mind-map $M$ is generated by removing unrelated nodes and edges.

Our work focuses on the former phase $D \rightarrow \mathbf{G}$. In the following sections, we construct the coreference graph to model the long-range relationships in a document. Then, based on the coreference graph, we employ a coreference graph encoder and a graph enhancement module to extract better representations, which can help optimize the relation graph. For the latter phase $\textbf{G} \rightarrow M$, two types of mind-maps are generated from the graph $\textbf{G}$ by pruning the extra edges. The strategy works recursively to determine the governing relationship of sentences. Please refer to \citet{hu2021efficient} for details of phase 2. 

\subsection{Coreference Graph}

\begin{figure}
    \includegraphics[width=0.48\textwidth]{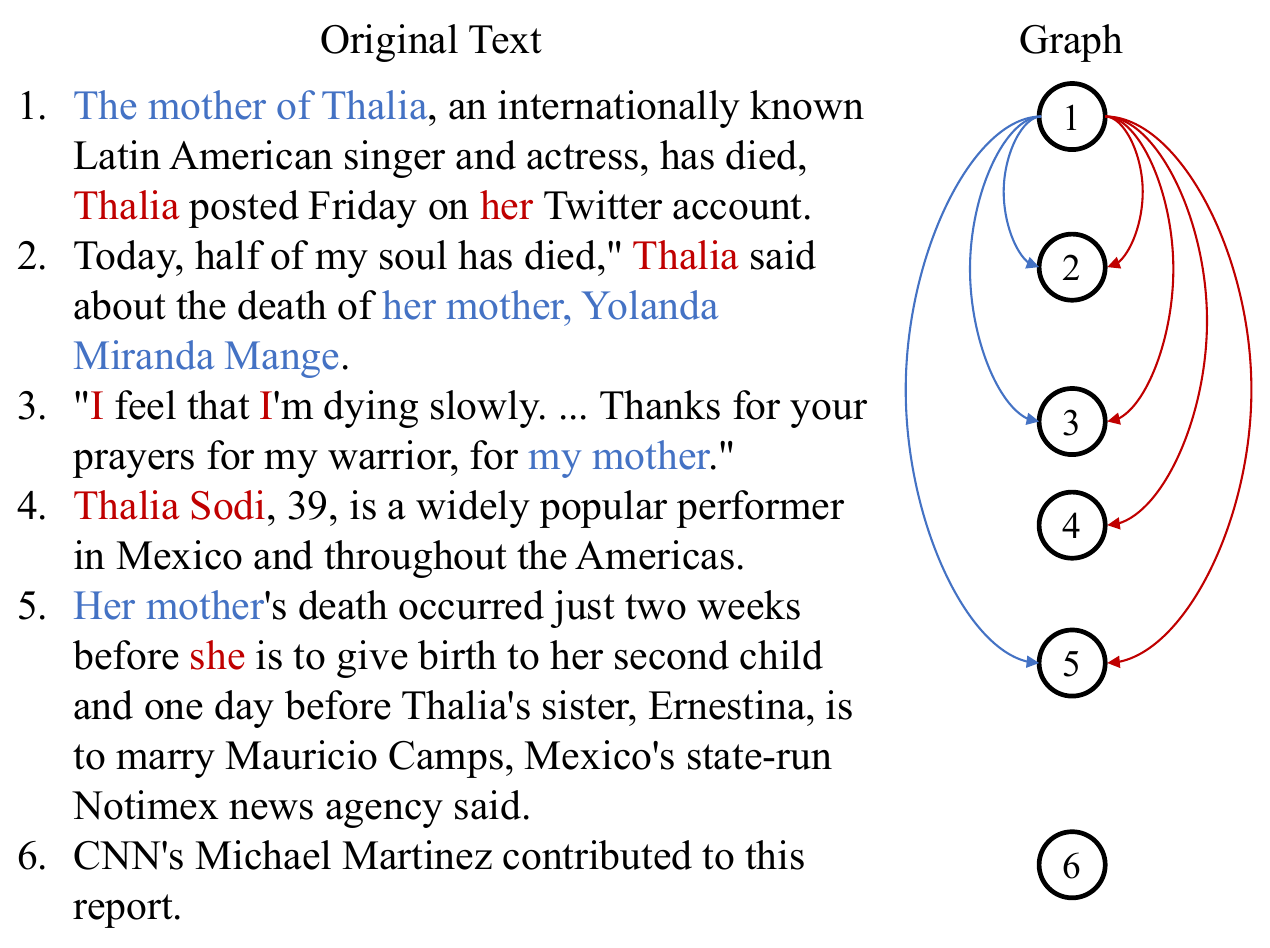}
\caption{Illustration of coreference graph construction, with the mentions of \emph{``Thalia''} 
 and \emph{``Thalia's mother''} highlighted as examples.}

\end{figure}

\citet{hu2021efficient} propose a sequence-to-graph method to generate graphs from sequentially encoded sentences. Although this approach contributes to preserving the sequential semantic information within articles, it shows a restricted capability to perceive the structural information. In addition, the model is less effective in modeling long-range semantic relationships. In order to address these two issues, we employ the coreference graph \cite{xu2020discourse} to model the documents. This approach introduces explicit graph structures to incorporate additional structural information and improves the perception of long-distance semantic relations.

Algorithm 1 describes the construction process of the coreference graph. In this work, we treat a sentence as the minimal unit. We first use Allennlp \cite{lee2018higher} to find all mentions that refer to the same entity in a document and obtain the coreference clusters $\mathcal{C}=\{c_i\}_{i=1}^{|\mathcal{C}|}$. Each cluster $c_i$ contains a number of mentions $\{M_{1}^i, M_{2}^i, ..., M_{m}^i\}\in{c_i}$. Next, for each cluster $c_i \in \mathcal{C}$, we match the mentions to the sentences sequentially in which they are located. Then, for each cluster, we select the sentence containing the first occurring mention as the root node and create an edge from the root node to the other sentence where the mention is located. The edges indicate the potential governing relation. Following this process of iterating through all the clusters can complete the construction of the final coreference graph.

Figure 3 provides an example, where \emph{``Thalia''} and \emph{``Thalia's mother''} are two entities to which various mentions refer. Edges are constructed from the first sentence node to the others. Note that the final graph does not contain duplicate edges.

\begin{algorithm}[tb]
\caption{Coreference Graph Generation}
\label{alg:algorithm}
\textbf{Input}: Coreference clusters $\mathcal{C}=\{c_i\}_{i=1}^{|\mathcal{C}|}$ \\
Initial coreference graph $\mathcal{G}\leftarrow\bm{0}$ 
\begin{algorithmic}[1] 
\FOR{$i=1$ to $|\mathcal{C}|$}
\STATE Match each mention $\{M_{1}^i, M_{2}^i, ..., M_{m}^i\} \in c_i$ to its origin sentences and obtain the index set $\mathcal{I}$.
\FOR{$j = 2$ to $|\mathcal{I}|$}
\STATE Create an edge $\mathcal{G}_{\mathcal{I}[1],\mathcal{I}[j]} = 1$
\ENDFOR
\ENDFOR
\STATE \textbf{return} the final coreference graph $\mathcal{G}$
\end{algorithmic}
\end{algorithm}


\subsection{Coreference-Guided Relation Graph Generation}

\subsubsection{Document Encoder}
In order to obtain the representations for each node, 
we firstly map the given sentence $s_i$ into an embedding sequence $\{\bm{e_i^1, e_i^2,..., e_i^L}\}$ through a pre-trained embedding matrix GloVE \cite{pennington2014glove}. Then we exploit a Bi-directional LSTM (BiLSTM) \cite{graves2013speech} to encode the embedding sequence into the hidden states $\{\bm{h_i^1, h_i^2,..., h_i^L}\}$. The vector representation for each sentence is computed by a max-pooling operation over the hidden states. 
\begin{equation}
    \bm{s_k}=\mathrm{max}({\bm{h_i^1,..., h_i^L}})
\end{equation}

Additionally, to model the sentence-level context, we encode the vector representations of all sentences $\{\bm{s_i}\}_{i=1}^N$ with another BiLSTM. We take the output $\mathbf{H}=\{\bm{h_1, h_2, ..., h_N}\}$ as the final representations.

\begin{figure}
\centering
    \includegraphics[width=0.48\textwidth]{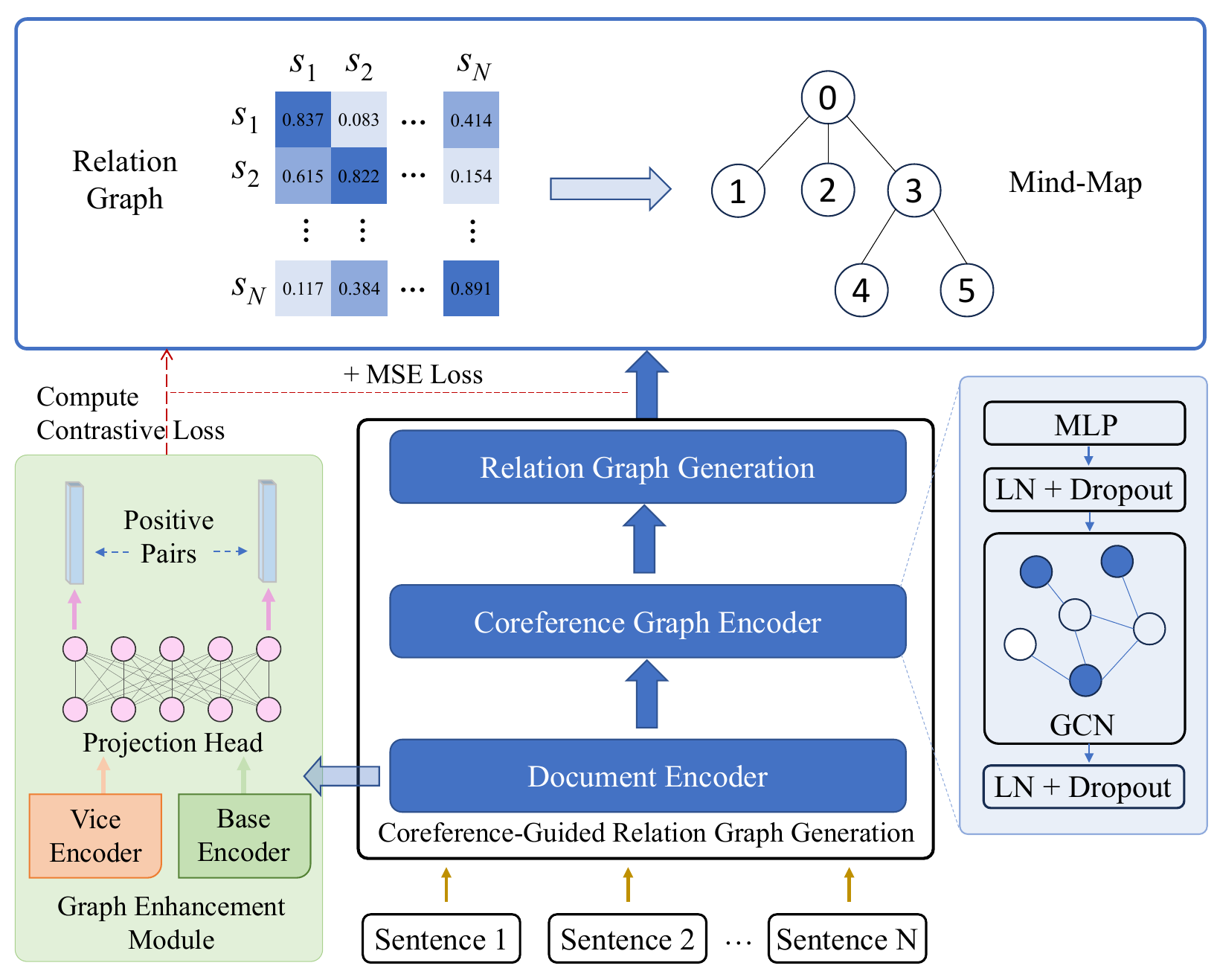}
\caption{The network architecture of the proposed approach, where the document $D$ is processed to obtain the relation graph, with the assistance of graph enhancement.}

\end{figure}

\subsubsection{Coreference Graph Encoder}
In the coreference graph $\mathcal{G=(V, E)}$, nodes $\mathcal{V}$ represent all sentences from the document and edges $\mathcal{E}$ represent the coreferential relationships between sentences. We utilize a GCN encoder to update the representations of all sentences \cite{xu2020discourse}. Figure 4 depicts the architecture design of the GCN network, which contains multiple layers of the same structure.

First, we define $\mathbf{H}\bm{^{(k)}}=\{\bm{h_1^{(k)}},...,\bm{h_N^{(k)}}\}$ as the input of the $k$-th layer in the model, which is also the output of the former layer (except the first layer). $\mathbf{H^{(1)}}=\mathbf{H}$, which is the output from the document encoder. The $k$-th layer is designed as follows:
\begin{equation}
    \bm{u_i^{(k)}}=\mathbf{W}\bm{_2^{(k)}}\mathrm{ReLU}(\mathbf{W}\bm{_1^{(k)}}\bm{h_i^{(k)}}+\bm{b_1^{(k)}})+\bm{b_2^{(k)}} 
\end{equation}
\begin{equation}
    \bm{v_i^{(k)}}=\mathrm{LN}(\bm{h_i^{(k)}}+\mathrm{Dropout}(\bm{u_i^{(k)}}))
\end{equation}
\begin{equation}
    \bm{w_i^{(k)}}=\mathrm{ReLU}(\sum_{j\in N_i}\frac{1}{\left | N_i \right | }\mathbf{W}\bm{_3^{(k)}} \bm{v_i^{(k)}} + \bm{b_3^{(k)}})
\end{equation}
\begin{equation}
    \bm{h_i^{(k+1)}}=\mathrm{LN}(\mathrm{Dropout}(\bm{w_i^{(k)}})+\bm{v_i^{(k)}})
\end{equation}

~\\
where $\mathrm{LN}(\cdot)$ is layer normalization, $N_i$ is the neighbourhood of the $i$-th node, and $\bm{h_i^{(k+1)}}$ is the output of the $i$-th node in the $k$-th layer. 

After multi-layer message propagation, we obtain the output of the last layer $\mathbf{R}=\{\bm{r_1, r_2, ..., r_N}\}$ as the final representation of all the sentences in the document.

\begin{table*}[]
\centering
\begin{tabular}{ll|cccc|cccc}
\toprule
\multicolumn{2}{c|}{\multirow{2}{*}{Models}} &
  \multicolumn{4}{c|}{SSM} &
  \multicolumn{4}{c}{KSM} \\
\multicolumn{2}{c|}{} & R-1   & R-2   & R-L   & Avg   & R-1   & R-2   & R-L   & Avg   \\ \midrule
\multirow{5}{*}{\begin{tabular}[c]{@{}c@{}}Baseline\\ Methods\end{tabular}} &
  Random &
  32.71 &
  23.51 &
  30.08 &
  28.77 &
  29.73 &
  26.50 &
  29.67 &
  28.63 \\
    & LexRank         & 34.53 & 25.04 & 31.79 & 30.45 & 31.04 & 27.75 & 31.00 & 29.93 \\
    & MRDMF           & 38.19 & 29.51 & 35.72 & 34.47 & 33.18 & 30.26 & 33.08 & 32.18 \\
    & DistilBERT      & 42.15 & 33.34 & 39.66 & 38.38 & 40.00 & 36.92 & 39.92 & 38.95 \\
    & EMGN            & 46.14 & 38.21 & 43.84 & 42.73 & 43.33 & 40.67 & 43.28 & 42.43 \\ \midrule
\multirow{4}{*}{\begin{tabular}[c]{@{}c@{}}Model\\ Variants\end{tabular}} &
  w/o CGE &
  26.44 &
  15.51 &
  23.26 &
  21.74 &
  26.11 &
  21.63 &
  26.05 &
  24.60 \\
    & w/o GEM         & 45.88 & 37.60 & 43.44 & 42.31 & 43.03 & 40.07 & 43.10 & 42.08 \\
    & w/o CGE\&GEM    & 44.99 & 37.02 & 42.71 & 41.57 & 42.56 & 39.58 & 42.50 & 41.55 \\
    \midrule
Full Model &
  CMGN &
  \textbf{47.73} &
  \textbf{39.75} &
  \textbf{45.33} &
  \textbf{44.27} &
  \textbf{44.65} &
  \textbf{41.77} &
  \textbf{44.59} &
  \textbf{43.67} \\ \bottomrule
\end{tabular}
\caption{Evaluation results of SSM and KSM in terms of R-1 (\%), R-2 (\%), R-L (\%) and the average score (\%). Model variants are further described in the ablation study section.}
\label{table:overall}
\end{table*}

\subsubsection{Relation Graph Generation}

We employ sequence-to-graph \cite{dozat2016deep, zhang2019amr}to process the final representations into a relation graph.  Concretely, we first compute the representations of all sentences when they are regarded as the start or end nodes in the edges separately. Then we calculate the governing scores with a bilinear operation. The process can be described as follows:
\begin{equation}
    \bm{r_i^{start}} = \mathrm{MLP}_1(\bm{r_i})
\end{equation}
\begin{equation}
    \bm{r_j^{end}} = \mathrm{MLP}_2(\bm{r_j})
\end{equation}
\begin{equation}
    \textbf{G}_{i,j} = \mathrm{\sigma}(\bm{r_i^{start}} \mathbf{W_4} \bm{r_j^{end}} + \bm{b_4})
\end{equation}
where $\mathrm{MLP}_1$ and $\mathrm{MLP}_2$ are two linear transformations, $\mathbf{W_4}$ is the parameter matrix, and $\bm{b_4}$ is the bias. $\sigma$ is the sigmoid operation, guaranteeing that each governing score is between 0 and 1. By calculating the scores for all pairs of sentences, we get the relation graph $\textbf{G}$.

\subsection{Graph Enhancement Module}

The graph structure information introduced in the previous section is not always reliable. Following \citet{xia2022simgrace}, we employ a graph enhancement module based on GCL to address this issue, which consists of the following three major components.

\subsubsection{Encoder Perturbation}
We first define a GNN encoder $f(\cdot;\bm{\theta})$ as the base encoder, and its perturbed version $f(\cdot;\bm{\theta'})$ as the vice encoder for contrasting. The method to perturb the encoder $f(\cdot;\bm{\theta})$ can be described mathematically as:
\begin{equation}
    \bm{\theta_l'}=\bm{\theta_l}+\eta \cdot \bm{\Delta \theta_l}
\end{equation}
\begin{equation}
    \bm{\Delta \theta_l} \sim \mathcal{N}(0, \sigma_l^2)
\end{equation}
where $\bm{\theta_l}$ and $\bm{\theta_l'}$ are the weight tensors of the $l$-th layer of the GNN encoder and its perturbed version respectively. $\eta$ is the hyperparameter used to regulate the perturbation. $\bm{\Delta \theta_l}$ is the perturbation term which samples from Gaussian distribution with zero mean and variance $\sigma_l^2$.

Then, we use the same coreference graph $\mathcal{G}$ and the output of document encoder $\mathbf{H}=\{\bm{h_1}, \bm{h_2}, ..., \bm{h_N}\}$ as the input of both GNN encoders, and obtain extracted representation:
\begin{equation}
    \bm{h}=f(\mathcal{G};\bm{\theta})
\end{equation}
\begin{equation}
    \bm{h'}=f(\mathcal{G};\bm{\theta'})
\end{equation}

\subsubsection{Projection Head}
A projection head is a non-linear transformation that can map representations to another linear space. This is beneficial in defining contrastive loss \cite{chen2020simple}. We employ a two-layer MLP to obtain:
\begin{equation}
\bm{z}=g(\bm{h})=\mathbf{W_6}\mathrm{ReLU}(\mathbf{W_5}\bm{h})
\end{equation}
\begin{equation}
    \bm{z'}=g(\bm{h'})=\mathbf{W_6}\mathrm{ReLU}(\mathbf{W_5}\bm{h'})
\end{equation}

\subsubsection{Contrastive Loss}
Following previous works \cite{oord2018representation, sohn2016improved, wu2018unsupervised, you2020graph, xia2022simgrace}, we use the normalized temperature-scaled cross-entropy loss (NT-Xent) as loss function, which benefits agreement between positive pairs. Concretely, we randomly sample a minibatch of $N$ examples and then feed them to the base encoder $f(\cdot;\bm{\theta})$ and its perturbed version $f(\cdot;\bm{\theta'})$. We can obtain a total of 2$N$ examples as output. We select negative examples from the other $N-1$ perturbed representations for each positive pair. Finally, for the $i$-th representation, the contrastive loss is defined as:
\begin{equation}
    l_i=-\log\frac{\exp (\mathrm{sim}(\bm{z_i}, \bm{z_i'})/\tau )}{{\textstyle \sum_{i'=1, i'\ne i}^{N}} \exp (\mathrm{sim}(\bm{z_i}, \bm{z_i'})/\tau )}
\end{equation}
where $\mathrm{sim}(\bm{z_i}, \bm{z_i'})=\bm{z^\top} \bm{z'}/\left\| \bm{z} \right\| \left \| \bm{z'} \right \|$ is the cosine similarity function, and $\tau$ is the temperature parameter. The final loss is computed across all positive pairs:
\begin{equation}
    \mathcal{L}_c = {\textstyle \sum_{i=1}^{N}} l_i
\end{equation}

\subsection{Training Objective}
In the coreference graph encoder module, our training goal is to fit the relation graph $\textbf{G}$ to the pseudo graph annotated by DistilBERT \cite{sanh2019distilbert}. The DistilBERT model is fine-tuned to automatically annotate a large number of relation labels rapidly\footnote{For details on fine-tuning, please refer to \citet{hu2021efficient}.}. The graph encoder module fits the pseudo graph $\textbf{Y}$ by a mean square error (MSE) loss.
\begin{equation}
    \mathcal{L}_g=\frac{1}{N^2} \sum_{i}^{} \sum_{j}^{}(\textbf{G}_{i, j} - \textbf{Y}_{i, j})^2
\end{equation}

Then we combine the final loss $\mathcal{L}_c$ and the training loss of coreference graph encoder $\mathcal{L}_g$ as an overall training objective to optimize the parameters $\theta$. 
\begin{equation}
    \mathcal{L}=\mathcal{L}_g+\lambda \mathcal{L}_c
\end{equation}
where $\lambda$ is the hyperparameter adjusting the influence of the graph enhancement module.

\section{Experiments}

\subsection{Dataset}

We use a human-annotated evaluation benchmark with 135 articles\footnote{\url{https://github.com/hmt2014/MindMap}}, 98,181 words, which are selected from CNN news articles \cite{hermann2015teaching, cheng2016neural} randomly. The benchmark consists of a testing set $\mathcal{D}_t$ and a validation set $\mathcal{D}_v$, with 120 and 15 articles respectively. We select 5000 articles from CNN news, where both the length and the number of sentences are no more than 50. We utilize the fine-tuned DistilBERT to annotate their pseudo graphs, which are then used for model training.

\subsection{Implementation Details}

\subsubsection{Software and Hardware}
We compare all the models in the same software and hardware environments, as follows:

\begin{itemize}
    \item System: Ubuntu 22.04.2; Python 3.7; PyTorch 1.12.1; DGL 1.0.2+cu116 (For the implementation of GNNs)
    \item CPU: Intel(R) Xeon(R) Gold 5218 CPU @ 2.30GHz
    \item GPU: NVIDIA Tesla V100S PCIe 32 GB
\end{itemize}

\subsubsection{Hyperparameter Settings}
We initialize the word embeddings with 50-dimension GloVE. We initialize all other parameters by sampling from a normal distribution with $\mathcal{N}$(0, 0.02). The hidden size of BiLSTM is set to be 25×2. The models are optimized by Adam \cite{kingma2014adam} with a learning rate of 1e-4. The batch size is 64. An early stop strategy is utilized during training if there is no performance improvement on the validation set $\mathcal{D}_v$ in 3 epochs, and the best model is saved for evaluating testing set $\mathcal{D}_t$. All the reported results are the average score of 5 runs.

For the coreference graph encoder, we employ a 2-layer model for training. For the graph enhancement module, we employ a 5-layer GIN \cite{xu2018powerful} model as the base encoder. We set $\eta$ to 0.2, which adjusts the magnitude of the perturbation of the base encoder, and $\lambda$ to 0.001, which adjusts the impact of contrastive loss.

\subsection{Compared Methods}
\subsubsection{Baselines Methods}
We validate our method by comparing its effectiveness with the following baseline methods.
\begin{itemize}
    \item \textbf{Random}: We randomly sample a graph $\textbf{G}$ for a document. Each governing score $\textbf{G}_{i,j}$ ranges from zero to one.
    \item \textbf{LexRank}: It computes the governing score of sentence pair by the cosine similarity of their TFIDF vectors. 
    \item \textbf{MRDMF} \cite{wei2019revealing}: It employs multi-hop self-attention and gated recurrence network to reveal the semantic relations via sentences, and then select the most salient sentences recursively to constitute the hierarchy.
    \item \textbf{DistilBERT} \cite{sanh2019distilbert}: It's a lighter version of BERT \cite{devlin2019bert} which helps annotate the pseudo graphs for training. 
    \item \textbf{EMGN} \cite{hu2021efficient}: This is the state-of-the-art mind-map generation work. It converts a document into a graph via sequence-to-graph and designs a graph refinement module based on reinforcement learning.
\end{itemize}

\subsubsection{Model Variants} The proposed method is further analyzed by changing individual components.

\begin{itemize}
    \item \textbf{w/o CGE}: The coreference graph encoder (CGE) is removed from the whole model.
    \item \textbf{w/o GEM}: The graph enhancement module (GEM) is removed from the whole model.
    \item \textbf{w/o CGE\&GEM}: Both graph encoder and graph enhancement module are removed from the whole model.
\end{itemize}

\begin{table}[]
\begin{tabular}{l|cccc}
\toprule
Models        & 0-shot & 1-shot & \multicolumn{1}{l}{2-shot} & \multicolumn{1}{l}{3-shot} \\ 
\midrule
ChatGPT       & 42.36  & 40.38  & 45.87                      & 39.96                      \\
ChatGPT w/ CoT & 43.97  & -      & -                          & -                          \\
LLaMA2-70B     & 33.12  & 32.10  & 35.62                      & -                          \\
ErnieBot-4.0   & 36.59  & 38.29  & 35.13                      & -                          \\ 
\midrule
CMGN          & \multicolumn{4}{c}{\textbf{52.41}}                                  \\ \bottomrule
\end{tabular}
\caption{Evaluation results of LLMs in terms of SSM.}
\end{table}

\begin{table}[]
\begin{tabular}{l|cccc}
\toprule
Models        & 0-shot & 1-shot & \multicolumn{1}{l}{2-shot} & \multicolumn{1}{l}{3-shot} \\ \midrule
ChatGPT       & 42.25  & 41.20  & 45.84                      & 41.32                      \\
ChatGPT w/ CoT & 42.48  & -      & -                          & -                          \\
LLaMA2-70B     & 33.57  & 33.94  & 36.27                      & -                          \\
ErnieBot-4.0   & 36.28  & 37.83  & 36.95                      & -                          \\ \midrule
CMGN          & \multicolumn{4}{c}{\textbf{51.26}}                                        \\ \bottomrule
\end{tabular}
\caption{Evaluation results of LLMs in terms of KSM.}
\end{table}

\subsection{Experimental Results}
\subsubsection{Overall Results}
The overall results are displayed in Table 1. It can be observed that CMGN achieves the best performance in both types of mind-map consistently, outperforming the state-of-the-art model EMGN by +1.54\% and +1.24\% on average of SSM and KSM, respectively. This indicates the effectiveness of CMGN. Moreover, though CMGN utilizes pseudo graph labels, it outperforms DistilBERT significantly. This observation highlights the successful utilization of pseudo labels by the proposed method, resulting in further enhancements in performance. 

\subsubsection{Ablation Study}
To validate the impact of individual components, several model variants are designed and the results are presented in Table 1. Firstly, it can be seen that w/o CGE causes performance to drop drastically. A possible reason is that w/o CGE and solely using contrastive learning may lead to negative effects on the representations of the document encoder. In this situation, the document encoder can not learn sequential information like EMGN but receive inaccurate graph representations from GEM. Secondly, we observe that w/o GEM leads to consistent degradation. This validates that our method can successfully enhance the graph representations. Moreover, the results of w/o CGE\&GEM further demonstrate the effectiveness of CMGN.


\subsubsection{Compared with LLMs}
The large language model (LLM) has proven its impressive skills in different language tasks. Thus, we try to test its potential in mind-map generation. However, using LLMs to generate a mind-map directly resulted in unstable outcomes, making automatic evaluation challenging. Thus, we make LLMs to generate a relation graph (phase 1), which is further fed into phase 2 as other approaches for evaluation. The prompts are shown below. 
\begin{tcolorbox}[colback=gray!10]
    \textbf{Prompt}: \textit{Given a document which contains N sentences, you have to build an N*N relation graph G for all pairs of sentences. Each element in G is the governing relationship score between two sentence. For instance, the element in [2, 1] indicates the intensity of sentence 2 governing sentence 1. Here are several notes: 1. Guarantee that each governing score is between 0 and 1. 2. Output the relation graph in the format of Python list. 3. Don't provide me the code. Generate the relation graph directly. Now I'm giving you the document. Are you ready?}
\end{tcolorbox}
 We select ChatGPT, LLaMA2-70B \cite{touvron2023llama}, and ErnieBot-4.0 for testing. In addition, we apply few-shot settings and the Chain-of-Thought strategy \cite{wei2022chain}. Due to the context length limitation, we only evaluate documents with $\le$25 sentences. The results are shown in Table 2 and 3. It can be observed that CMGN significantly outperforms all LLMs. This shows that LLMs are universal AI systems, which might lack specialty in particular tasks. Even though, LLMs are still powerful since it outperforms supervised methods, such as LexRank and MRDMF.

\begin{table}[]
\centering
\begin{tabular}{l|cc|cc}
\toprule
\multirow{2}{*}{Models} & \multicolumn{2}{c|}{SSM Avg}                      & \multicolumn{2}{c}{KSM Avg}                           \\
                        & $\le$25                  & \textgreater{}25           & $\le$25                       & \textgreater{}25          \\ \midrule
Random                  & 31.94                & 26.63                      & 32.53                     & 26.22                     \\
LexRank                 & 33.54                & 28.07                      & 33.99                     & 27.28                     \\
MRDMF                   & 39.83                & 30.58                      & 36.76                     & 28.69                     \\
DistilBERT              & 44.36                & 34.34                      & 44.27                     & 34.89                     \\
EMGN                    & 50.23                & 37.38                      & 49.48                     & 37.44                     \\ \midrule
\multicolumn{1}{l|}{CMGN}   & \textbf{52.41}                & \multicolumn{1}{l|}{\textbf{38.46}} & \multicolumn{1}{l}{\textbf{51.26}} & \multicolumn{1}{l}{\textbf{38.20}} \\ \bottomrule
\end{tabular}
\caption{Evaluation results by splitting the testing set with the sentence number in a document. Among
all 120 files in $\mathcal{D}_t$, there are 50 files with $\le$ 25 sentences, and 70 files $>$ 25.}
\end{table}

\begin{table}[]
\centering
\begin{tabular}{l|cc}
\toprule
Models           & SSM Avg        & KSM Avg        \\ \midrule
CMGN(BERT)       & 39.22          & 38.82          \\
CMGN(DistilBERT) & 38.12          & 38.92          \\
CMGN(RoBERTa)    & 38.27          & 39.54          \\
CMGN(Sentence-BERT)      & 39.09          & 39.53          \\ \midrule
CMGN             & \textbf{43.71} & \textbf{43.20} \\ \bottomrule
\end{tabular}
\caption{Evaluation results by different encoders.}
\end{table}

\subsubsection{Effects of the Document Length}
To analyze performance across varying document lengths, the testing set $\mathcal{D}_t$ is split into two parts based on document length. Table 4 displays the evaluation results. Our model outperforms all baselines, especially on the dataset with longer documents. Since it's a more challenging task to generate accurate mind-map for longer documents, the results indicate the effectiveness of our model in modeling long-range semantic relationships.

\subsubsection{Effects of Transformer-based Encoders}
We try employing more transformer-based pretrained models, including BERT \cite{devlin2019bert}, DistilBERT \cite{sanh2019distilbert}, RoBERTa \cite{liu2019roberta} and Sentence-BERT \cite{reimers2019sentence} to derive a sentence embedding. The results are shown in Table 5. It show that transformer-based embeddings do not contribute to the performances. A possible reason is that the self-attention mechanism in transformer architecture may make word representations in a sentence difficult to distinguish. On the contrary, GloVe is trained using word co-occurrence statistics, relying on the concept that words frequently appearing together share semantic relationships. This approach captures global, long-range word relationships, making it useful for tasks like mind-map generation that demand understanding both sentence and keyword semantics.

\subsection{Further Analysis}

\subsubsection{Inference Time}
We compare the inference time of the validation set $\mathcal{D}_v$ and testing set $\mathcal{D}_t$ to verify the efficiency of our method. Table 6 demonstrates the time spent to convert a document to a relation graph (phase 1), while all the methods share the follow-up process (phase 2). The batch size of MRDMF is set to 256 sentence pairs, and the batch size of EMGN and CMGN-related methods is 32 documents. The result shows that our method is slightly more efficient than EMGN, while significantly reducing the computational complexity compared to other methods. This proves that our model improves performance without losing efficiency.

\subsubsection{Hyperparameter Study}
In the graph enhancement module, $\eta$ determines the extent of perturbation of the vice encoder. $\lambda$ balances the effect of the graph enhancement module. Figure 5 shows the performance curves of different $\eta$ and $\lambda$ values. We find a strong correlation between the performance of the model and the parameter $\lambda$, while it is not sensitive to variations in parameter $\eta$.

\begin{table}[]
\centering
\begin{tabular}{l|c|c}
\toprule
Models     & $\mathcal{D}_t$   & $\mathcal{D}_v$  \\ \midrule
LexRank    & 349.21  & 95.24  \\
MRDMF      & 467.07  & 62.49  \\
DistilBERT & 1219.51 & 201.04 \\
EMGN       & 10.27   & 1.42   \\ \midrule
CMGN       & \textbf{4.09} & \textbf{1.24}   \\ \bottomrule
\end{tabular}
\caption{Inference time for each method (second).}
\end{table}

\begin{figure}
    \centering
    \subfigure[$\lambda$]{
        \includegraphics[width=0.22\textwidth]{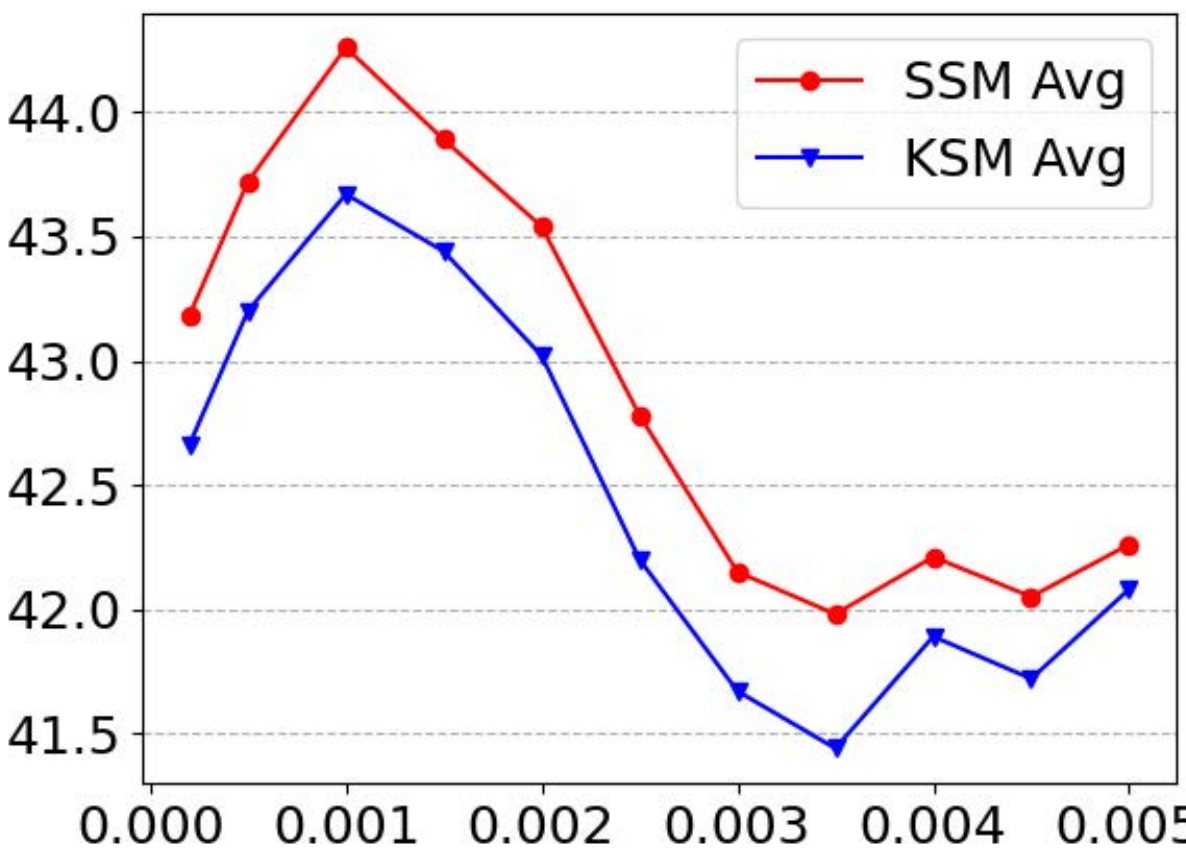}
    }
    \subfigure[$\eta$]{
        \includegraphics[width=0.22\textwidth]{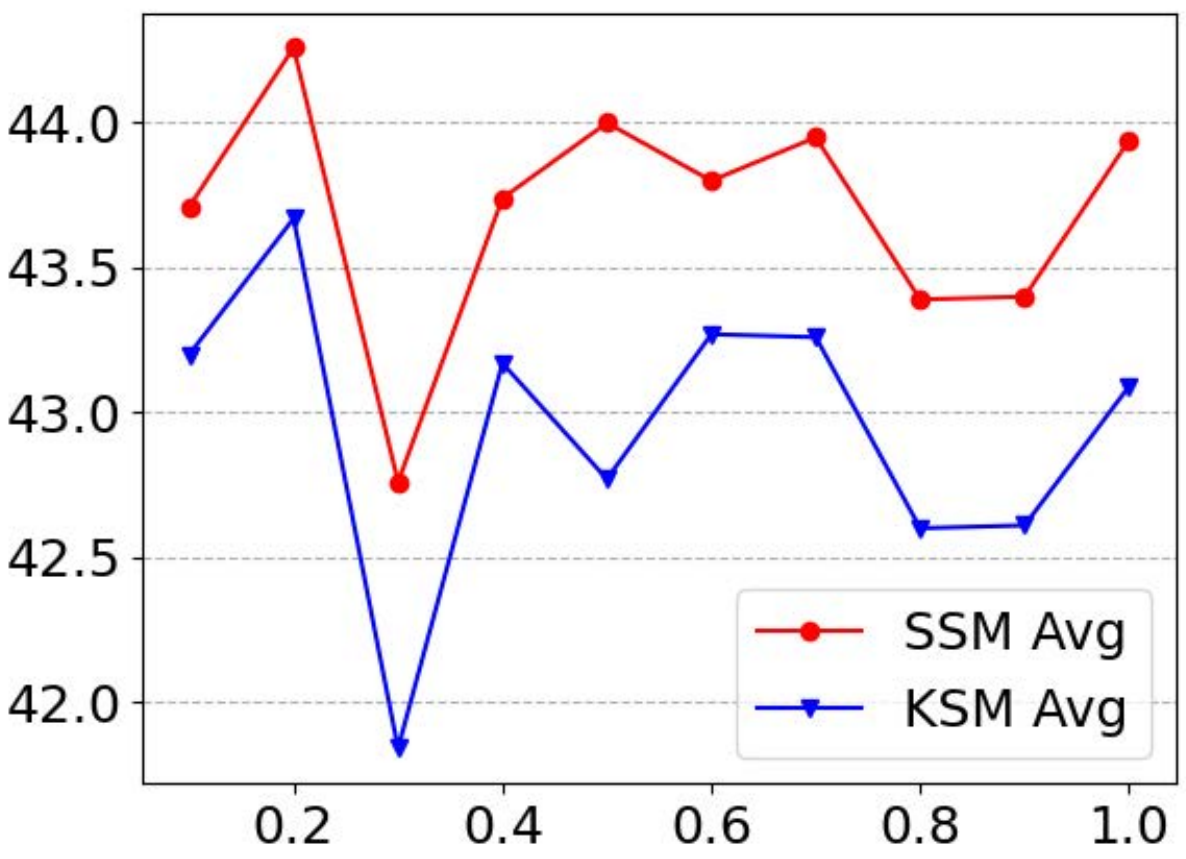}
    }
  \caption{Hyperparameter study for SSM and KSM.}
\end{figure}

\begin{figure}[t]
    \includegraphics[width=0.48\textwidth]{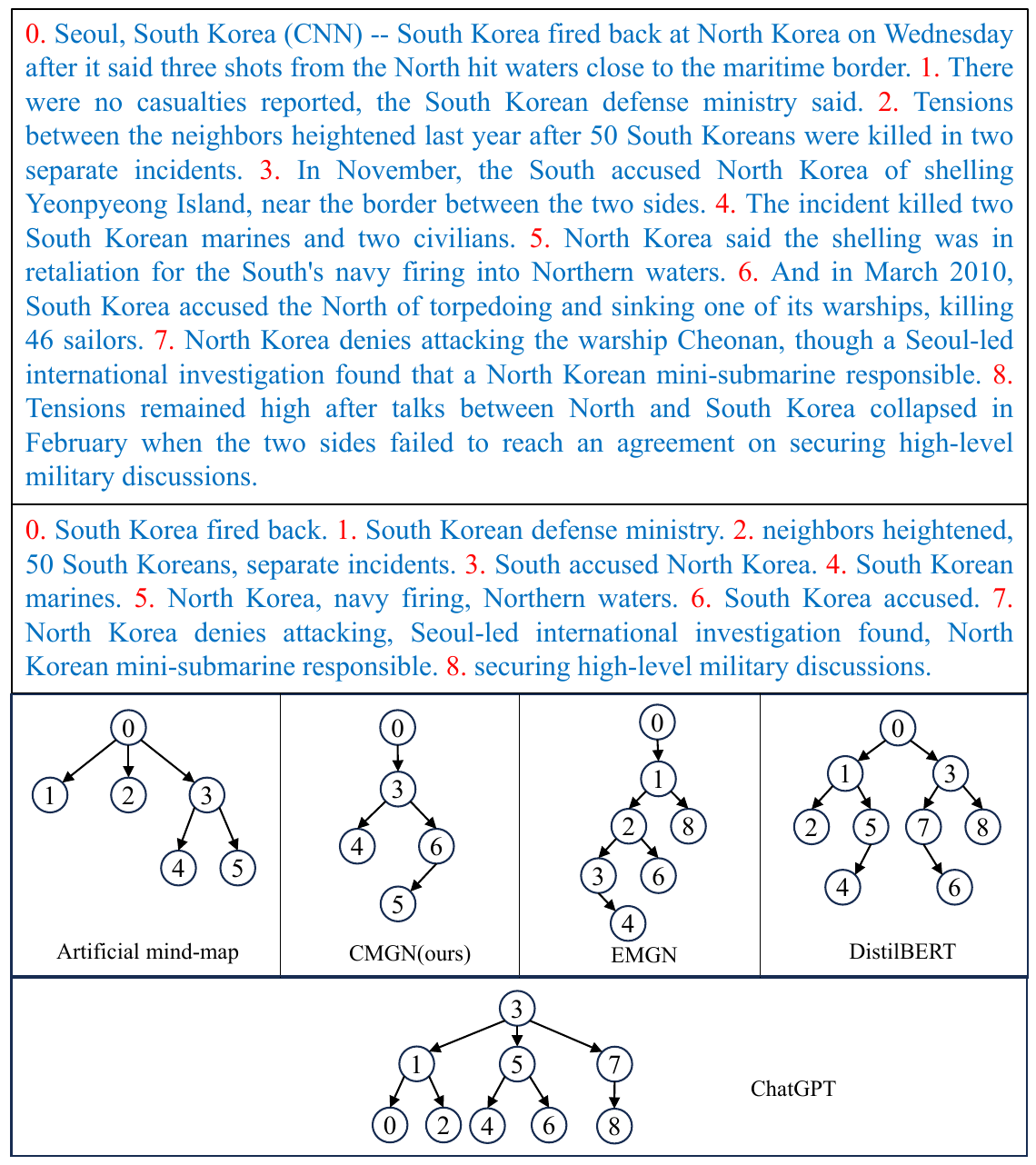}
\caption{Case study for SSM and KSM.}

\end{figure}

\subsubsection{Case Study}
Figure 6 depicts the mind-maps generated by different models, as well as the manually annotated mind-map\footnote{The original text can be found at \url{http://edition.cnn.com/2011/WORLD/asiapcf/08/10/koreas.unrest/index.html}}. Firstly, it can be observed that our model better demonstrates the governing relations of sentences 3-5 in general, which are the core of the document. In contrast, the mind-map by EMGN overlooks sentence 5, and the mind-map by DistilBERT places sentences 4-5 under the wrong governor. Furthermore, compared with other methods, ChatGPT selects an inaccurate root node, while other models commonly select sentence 0 as the root node. This reveals that LLMs like ChatGPT may have limitations on particular tasks. Small models are still essential in future research. In conclusion, the case study further illustrates that our model can reveal the hierarchical structure and semantic relationships of articles more accurately and concisely.

\section{Conclusion}
We propose an effective mind-map generation network based on coreference graph and coreference-guided relation graph generation. To better utilize the information of the coreference graph, we employ a graph enhancement module in a contrastive learning manner. The experiment results demonstrate the effectiveness of our method, while enjoying efficiency. The case study further proves that the mind-map of our method is better organized than other methods.

\bibliography{aaai24}

\begin{thebibliography}{40}
\providecommand{\natexlab}[1]{#1}

\bibitem[{Brucks and Schommer(2008)}]{brucks2008assembling}
Brucks, C.; and Schommer, C. 2008.
\newblock Assembling actor-based mind-maps from text stream.
\newblock \emph{arXiv preprint arXiv:0810.4616}.

\bibitem[{Buzan and Buzan(2006)}]{buzan2006mind}
Buzan, T.; and Buzan, B. 2006.
\newblock \emph{The mind map book}.
\newblock Pearson Education.

\bibitem[{Chen et~al.(2020)Chen, Kornblith, Norouzi, and Hinton}]{chen2020simple}
Chen, T.; Kornblith, S.; Norouzi, M.; and Hinton, G. 2020.
\newblock A simple framework for contrastive learning of visual representations.
\newblock In \emph{International conference on machine learning (ICML)}, 1597--1607. PMLR.

\bibitem[{Cheng and Lapata(2016)}]{cheng2016neural}
Cheng, J.; and Lapata, M. 2016.
\newblock Neural Summarization by Extracting Sentences and Words.
\newblock In \emph{Proceedings of the 54th Annual Meeting of the Association for Computational Linguistics (ACL)}, 484--494.

\bibitem[{Devlin et~al.(2019)Devlin, Chang, Lee, and Toutanova}]{devlin2019bert}
Devlin, J.; Chang, M.-W.; Lee, K.; and Toutanova, K. 2019.
\newblock BERT: Pre-training of Deep Bidirectional Transformers for Language Understanding.
\newblock In \emph{Proceedings of the 2019 Conference of the North {A}merican Chapter of the Association for Computational Linguistics: Human Language Technologies (NAACL-HLT)}, 4171--4186.

\bibitem[{Dhindsa and Roger~Anderson(2011)}]{dhindsa2011constructivist}
Dhindsa, H.~S.; and Roger~Anderson, O. 2011.
\newblock Constructivist-visual mind map teaching approach and the quality of students’ cognitive structures.
\newblock \emph{Journal of Science Education and Technology}, 20: 186--200.

\bibitem[{Dozat and Manning(2016)}]{dozat2016deep}
Dozat, T.; and Manning, C.~D. 2016.
\newblock Deep Biaffine Attention for Neural Dependency Parsing.
\newblock In \emph{International Conference on Learning Representations (ICLR)}, 1--8.

\bibitem[{Elhoseiny and Elgammal(2012)}]{elhoseiny2012english2mindmap}
Elhoseiny, M.; and Elgammal, A. 2012.
\newblock English2mindmap: An automated system for mindmap generation from english text.
\newblock In \emph{2012 IEEE International Symposium on Multimedia}, 326--331. IEEE.

\bibitem[{Erkan and Radev(2004)}]{erkan2004lexrank}
Erkan, G.; and Radev, D.~R. 2004.
\newblock Lexrank: Graph-based lexical centrality as salience in text summarization.
\newblock \emph{Journal of artificial intelligence research}, 22: 457--479.

\bibitem[{Graves, Mohamed, and Hinton(2013)}]{graves2013speech}
Graves, A.; Mohamed, A.-r.; and Hinton, G. 2013.
\newblock Speech recognition with deep recurrent neural networks.
\newblock In \emph{2013 IEEE international conference on acoustics, speech and signal processing (ICASSP)}, 6645--6649. Ieee.

\bibitem[{Hermann et~al.(2015)Hermann, Kocisky, Grefenstette, Espeholt, Kay, Suleyman, and Blunsom}]{hermann2015teaching}
Hermann, K.~M.; Kocisky, T.; Grefenstette, E.; Espeholt, L.; Kay, W.; Suleyman, M.; and Blunsom, P. 2015.
\newblock Teaching machines to read and comprehend.
\newblock \emph{Advances in neural information processing systems (NeurIPS)}, 28.

\bibitem[{Hu et~al.(2022)Hu, Li, Chen, Li, Wan, and Chang}]{hu-etal-2022-graph}
Hu, J.; Li, Z.; Chen, Z.; Li, Z.; Wan, X.; and Chang, T.-H. 2022.
\newblock Graph Enhanced Contrastive Learning for Radiology Findings Summarization.
\newblock In \emph{Proceedings of the 60th Annual Meeting of the Association for Computational Linguistics (ACL)}, 4677--4688.

\bibitem[{Hu et~al.(2021)Hu, Guo, Zhao, Gao, and Su}]{hu2021efficient}
Hu, M.; Guo, H.; Zhao, S.; Gao, H.; and Su, Z. 2021.
\newblock Efficient Mind-Map Generation via Sequence-to-Graph and Reinforced Graph Refinement.
\newblock In \emph{Proceedings of the 2021 Conference on Empirical Methods in Natural Language Processing (EMNLP)}, 8130--8141.

\bibitem[{Kingma and Ba(2015)}]{kingma2014adam}
Kingma, D.~P.; and Ba, J. 2015.
\newblock Adam: A method for stochastic optimization.
\newblock In \emph{The 3rd International Conference on Learning Representations (ICLR)}, 1--15.

\bibitem[{Kudeli{\'c}, Konecki, and Malekovi{\'c}(2011)}]{kudelic2011mind}
Kudeli{\'c}, R.; Konecki, M.; and Malekovi{\'c}, M. 2011.
\newblock Mind map generator software model with text mining algorithm.

\bibitem[{Lee, He, and Zettlemoyer(2018)}]{lee2018higher}
Lee, K.; He, L.; and Zettlemoyer, L. 2018.
\newblock Higher-Order Coreference Resolution with Coarse-to-Fine Inference.
\newblock In \emph{Proceedings of the 2018 Conference of the North American Chapter of the Association for Computational Linguistics: Human Language Technologies (NAACL-HLT)}, 687--692.

\bibitem[{Lin(2004)}]{lin2004rouge}
Lin, C.-Y. 2004.
\newblock Rouge: A package for automatic evaluation of summaries.
\newblock In \emph{Text summarization branches out}, 74--81.

\bibitem[{Liu, Hughes, and Yang(2021)}]{liu2021unsupervised}
Liu, J.; Hughes, D.~J.; and Yang, Y. 2021.
\newblock Unsupervised extractive text summarization with distance-augmented sentence graphs.
\newblock In \emph{Proceedings of the 44th International ACM SIGIR Conference on Research and Development in Information Retrieval (SIGIR)}, 2313--2317.

\bibitem[{Liu et~al.(2019)Liu, Ott, Goyal, Du, Joshi, Chen, Levy, Lewis, Zettlemoyer, and Stoyanov}]{liu2019roberta}
Liu, Y.; Ott, M.; Goyal, N.; Du, J.; Joshi, M.; Chen, D.; Levy, O.; Lewis, M.; Zettlemoyer, L.; and Stoyanov, V. 2019.
\newblock Roberta: A robustly optimized bert pretraining approach.
\newblock \emph{arXiv preprint arXiv:1907.11692}.

\bibitem[{Oord, Li, and Vinyals(2018)}]{oord2018representation}
Oord, A. v.~d.; Li, Y.; and Vinyals, O. 2018.
\newblock Representation learning with contrastive predictive coding.
\newblock \emph{arXiv preprint arXiv:1807.03748}.

\bibitem[{Parveen, Ramsl, and Strube(2015)}]{parveen2015topical}
Parveen, D.; Ramsl, H.-M.; and Strube, M. 2015.
\newblock Topical coherence for graph-based extractive summarization.
\newblock In \emph{Proceedings of the 2015 conference on empirical methods in natural language processing (EMNLP)}, 1949--1954.

\bibitem[{Parveen and Strube(2015)}]{parveen2015integrating}
Parveen, D.; and Strube, M. 2015.
\newblock Integrating importance, non-redundancy and coherence in graph-based extractive summarization.
\newblock In \emph{Proceedings of the 24th International Conference on Artificial Intelligence (IJCAI)}, 1298--1304.

\bibitem[{Pennington, Socher, and Manning(2014)}]{pennington2014glove}
Pennington, J.; Socher, R.; and Manning, C.~D. 2014.
\newblock Glove: Global vectors for word representation.
\newblock In \emph{Proceedings of the 2014 conference on empirical methods in natural language processing (EMNLP)}, 1532--1543.

\bibitem[{Reimers and Gurevych(2019)}]{reimers2019sentence}
Reimers, N.; and Gurevych, I. 2019.
\newblock Sentence-BERT: Sentence Embeddings using Siamese BERT-Networks.
\newblock In \emph{Proceedings of the 2019 Conference on Empirical Methods in Natural Language Processing and the 9th International Joint Conference on Natural Language Processing (EMNLP-IJCNLP)}, 3982--3992.

\bibitem[{Rothenberger et~al.(2008)Rothenberger, Oez, Tahirovic, and Schommer}]{rothenberger2008figuring}
Rothenberger, T.; Oez, S.; Tahirovic, E.; and Schommer, C. 2008.
\newblock Figuring out Actors in Text Streams: Using Collocations to establish Incremental Mind-maps.
\newblock \emph{arXiv preprint arXiv:0803.2856}.

\bibitem[{Sanh et~al.(2019)Sanh, Debut, Chaumond, and Wolf}]{sanh2019distilbert}
Sanh, V.; Debut, L.; Chaumond, J.; and Wolf, T. 2019.
\newblock DistilBERT, a distilled version of BERT: smaller, faster, cheaper and lighter.
\newblock In \emph{The 5th Workshop on Energy Efficient Machine Learning and Cognitive Computing of NeurIPS}.

\bibitem[{Sohn(2016)}]{sohn2016improved}
Sohn, K. 2016.
\newblock Improved deep metric learning with multi-class n-pair loss objective.
\newblock \emph{Advances in neural information processing systems (NeurIPS)}, 29.

\bibitem[{Sun et~al.(2022)Sun, Harit, Cristea, Yu, Shi, and Al~Moubayed}]{sun2022contrastive}
Sun, Z.; Harit, A.; Cristea, A.~I.; Yu, J.; Shi, L.; and Al~Moubayed, N. 2022.
\newblock Contrastive learning with heterogeneous graph attention networks on short text classification.
\newblock In \emph{2022 International Joint Conference on Neural Networks (IJCNN)}, 1--6. IEEE.

\bibitem[{Touvron et~al.(2023)Touvron, Martin, Stone, Albert, Almahairi, Babaei, Bashlykov, Batra, Bhargava, Bhosale, Bikel, Blecher, Ferrer, Chen, Cucurull, Esiobu, Fernandes, Fu, Fu, Fuller, Gao, Goswami, Goyal, Hartshorn, Hosseini, Hou, Inan, Kardas, Kerkez, Khabsa, Kloumann, Korenev, Koura, Lachaux, Lavril, Lee, Liskovich, Lu, Mao, Martinet, Mihaylov, Mishra, Molybog, Nie, Poulton, Reizenstein, Rungta, Saladi, Schelten, Silva, Smith, Subramanian, Tan, Tang, Taylor, Williams, Kuan, Xu, Yan, Zarov, Zhang, Fan, Kambadur, Narang, Rodriguez, Stojnic, Edunov, and Scialom}]{touvron2023llama}
Touvron, H.; Martin, L.; Stone, K.; Albert, P.; Almahairi, A.; Babaei, Y.; Bashlykov, N.; Batra, S.; Bhargava, P.; Bhosale, S.; Bikel, D.; Blecher, L.; Ferrer, C.~C.; Chen, M.; Cucurull, G.; Esiobu, D.; Fernandes, J.; Fu, J.; Fu, W.; Fuller, B.; Gao, C.; Goswami, V.; Goyal, N.; Hartshorn, A.; Hosseini, S.; Hou, R.; Inan, H.; Kardas, M.; Kerkez, V.; Khabsa, M.; Kloumann, I.; Korenev, A.; Koura, P.~S.; Lachaux, M.-A.; Lavril, T.; Lee, J.; Liskovich, D.; Lu, Y.; Mao, Y.; Martinet, X.; Mihaylov, T.; Mishra, P.; Molybog, I.; Nie, Y.; Poulton, A.; Reizenstein, J.; Rungta, R.; Saladi, K.; Schelten, A.; Silva, R.; Smith, E.~M.; Subramanian, R.; Tan, X.~E.; Tang, B.; Taylor, R.; Williams, A.; Kuan, J.~X.; Xu, P.; Yan, Z.; Zarov, I.; Zhang, Y.; Fan, A.; Kambadur, M.; Narang, S.; Rodriguez, A.; Stojnic, R.; Edunov, S.; and Scialom, T. 2023.
\newblock Llama 2: Open Foundation and Fine-Tuned Chat Models.
\newblock arXiv:2307.09288.

\bibitem[{Wang et~al.(2020)Wang, Liu, Zheng, Qiu, and Huang}]{wang2020heterogeneous}
Wang, D.; Liu, P.; Zheng, Y.; Qiu, X.; and Huang, X.-J. 2020.
\newblock Heterogeneous Graph Neural Networks for Extractive Document Summarization.
\newblock In \emph{Proceedings of the 58th Annual Meeting of the Association for Computational Linguistics (ACL)}, 6209--6219.

\bibitem[{Wei et~al.(2022)Wei, Wang, Schuurmans, Bosma, Xia, Chi, Le, Zhou et~al.}]{wei2022chain}
Wei, J.; Wang, X.; Schuurmans, D.; Bosma, M.; Xia, F.; Chi, E.; Le, Q.~V.; Zhou, D.; et~al. 2022.
\newblock Chain-of-thought prompting elicits reasoning in large language models.
\newblock \emph{Advances in Neural Information Processing Systems (NeurIPS)}, 35: 24824--24837.

\bibitem[{Wei et~al.(2019)Wei, Guo, Wei, and Su}]{wei2019revealing}
Wei, Y.; Guo, H.; Wei, J.-M.; and Su, Z. 2019.
\newblock Revealing Semantic Structures of Texts: Multi-grained Framework for Automatic Mind-map Generation.
\newblock In \emph{International Joint Conference on Artificial Intelligence (IJCAI)}, 5247--5254.

\bibitem[{Wu et~al.(2018)Wu, Xiong, Yu, and Lin}]{wu2018unsupervised}
Wu, Z.; Xiong, Y.; Yu, S.~X.; and Lin, D. 2018.
\newblock Unsupervised feature learning via non-parametric instance discrimination.
\newblock In \emph{Proceedings of the IEEE conference on computer vision and pattern recognition (CVPR)}, 3733--3742.

\bibitem[{Xia et~al.(2022)Xia, Wu, Chen, Hu, and Li}]{xia2022simgrace}
Xia, J.; Wu, L.; Chen, J.; Hu, B.; and Li, S.~Z. 2022.
\newblock Simgrace: A simple framework for graph contrastive learning without data augmentation.
\newblock In \emph{Proceedings of the ACM Web Conference 2022 (WWW)}, 1070--1079.

\bibitem[{Xu et~al.(2020)Xu, Gan, Cheng, and Liu}]{xu2020discourse}
Xu, J.; Gan, Z.; Cheng, Y.; and Liu, J. 2020.
\newblock Discourse-Aware Neural Extractive Text Summarization.
\newblock In \emph{Proceedings of the 58th Annual Meeting of the Association for Computational Linguistics (ACL)}, 5021--5031.

\bibitem[{Xu et~al.(2018)Xu, Hu, Leskovec, and Jegelka}]{xu2018powerful}
Xu, K.; Hu, W.; Leskovec, J.; and Jegelka, S. 2018.
\newblock How Powerful are Graph Neural Networks?
\newblock In \emph{International Conference on Learning Representations (ICLR)}, 1--17.

\bibitem[{Xu et~al.(2021)Xu, Chen, Ma, Huang, and Xiang}]{xu2021contrastive}
Xu, P.; Chen, X.; Ma, X.; Huang, Z.; and Xiang, B. 2021.
\newblock Contrastive Document Representation Learning with Graph Attention Networks.
\newblock In \emph{Proceedings of the 2021 conference on empirical methods in natural language processing (EMNLP Findings)}, 3874--3884.

\bibitem[{You et~al.(2020)You, Chen, Sui, Chen, Wang, and Shen}]{you2020graph}
You, Y.; Chen, T.; Sui, Y.; Chen, T.; Wang, Z.; and Shen, Y. 2020.
\newblock Graph contrastive learning with augmentations.
\newblock \emph{Advances in neural information processing systems (NeurIPS)}, 33: 5812--5823.

\bibitem[{Zhang et~al.(2019)Zhang, Ma, Duh, and Van~Durme}]{zhang2019amr}
Zhang, S.; Ma, X.; Duh, K.; and Van~Durme, B. 2019.
\newblock AMR Parsing as Sequence-to-Graph Transduction.
\newblock In \emph{Proceedings of the 57th Annual Meeting of the Association for Computational Linguistics (ACL)}, 80--94.

\bibitem[{Zhong et~al.(2020)Zhong, Liu, Chen, Wang, Qiu, and Huang}]{zhong2020extractive}
Zhong, M.; Liu, P.; Chen, Y.; Wang, D.; Qiu, X.; and Huang, X. 2020.
\newblock Extractive summarization as text matching.
\newblock In \emph{Proceedings of the 58th Annual Meeting of the Association for Computational Linguistics (ACL)}, 6197--6208.

\end{thebibliography}

\end{document}